# PST: Plant segmentation transformer for 3D point clouds of rapeseed plants at the podding stage


Ruiming Du [a, b], Zhihong Ma [a, b], Pengyao Xie [a, b], Yong He [a, b], Haiyan Cen [a, b]*

[a] *College of Biosystems Engineering and Food Science, and State Key Laboratory of Modern Optical Instrumentation, Zhejiang University, Hangzhou 310058, P.R. China*

[b] *Key Laboratory of Spectroscopy Sensing, Ministry of Agriculture and Rural Affairs, Hangzhou 310058, P.R. China*


## Abstract:


Segmentation of plant point clouds to obtain high-precise morphological traits is essential for plant phenotyping. Although the fast development of deep learning has boosted much research on segmentation of plant point clouds, previous studies mainly focus on the hard voxelization-based or down-sampling-based methods, which are limited to segmenting simple plant organs. Segmentation of complex plant point clouds with a high spatial resolution still remains challenging. In this study, we proposed a deep learning network plant segmentation transformer (PST) to achieve the semantic segmentation of rapeseed plants point clouds acquired by handheld laser scanning (HLS) with the high spatial resolution, which can characterize the tiny siliques as the main traits targeted. PST is composed of: (i) a dynamic voxel feature encoder (DVFE) to aggregate the point features with the raw spatial resolution; (ii) the dual window sets attention blocks to capture the contextual information; and (iii) a dense feature propagation module to obtain the final


dense point feature map. We then integrated PST with an instance segmentation head in the point grouping network (PointGroup) and developed PST-PointGroup (PG) to achieve the instance segmentation of the siliques. The results proved that PST and PST-PG achieved superior performance in semantic and instance segmentation tasks. For the semantic segmentation, the mean IoU, mean Precision, mean Recall, mean F1-score, and overall accuracy of PST were 93.96%, 97.29%, 96.52%, 96.88%, and 97.07%, achieving an improvement of 7.62, 3.28, 4.8, 4.25, and 3.88 percentage points compared to the second-best state-of-the-art network position adaptive convolution (PAConv). For instance segmentation, PST-PG reached 89.51%, 89.85%, 88.83% and 82.53% in mCov, mWCov, $mPerc_{90}$, and $mRec_{90}$, achieving an improvement of 2.93, 2.21, 1.99, and 5.9 percentage points compared to the original instance segmentation network PointGroup. This study extends the phenotyping of rapeseed plants in an end-to-end way and proves that the deep-learning-based point cloud segmentation method has a great potential for resolving dense plant point clouds with complex morphological traits.

**Keywords**: 3D deep learning; Point cloud segmentation; Handheld laser scanning; Plant phenotyping



# 1. Introduction

Phenotyping of plant morphological traits is essential in plant breeding and precision farm management (Tran et al., 2017). The rapeseed plant, with superior nutritional quality, is regarded



as one of the most important oil industrial crops and is cultivated worldwide (Friedt et al., 2007).

The siliques are the main contributions to the seed yield and quality of rapeseed plants, and silique

phenotypes are widely leveraged for yield estimation. Specifically, the growth locations of siliques

(Oleksy et al., 2018), mature degree (Wang et al., 2016), total numbers, and length (Wolko et al.,

2019) of siliques are highly related to the productivity and the oil quality of rapeseed plants.

Traditional methods for plant phenotyping rely on labor-intensive and destructive manual

measurement (Tanksley, 2004), while the fast development of optical sensors and imaging

technologies provides an opportunity for high-throughput approaches. The prerequisite of these

approaches to give reliable yield estimation lies in the precise segmentation of plant organs.

However, the structure of a rapeseed plant at the podding stage is complex, where plenty of tiny

siliques scatter and overlap within a plant, making it difficult to recognize them. The high-precise

segmentation of rapeseed plants, therefore, still remains a challenge.

Among current high-throughput pipelines, the most widely used two-dimensional (2D) image-

based approaches for plant organ segmentation are restricted by the illumination variation or organ

overlaps (Li et al., 2020). In contrast, accurate 3D data acquired from various sensors, such as

structured light, time-of-flight (TOF) cameras, and laser scanners, can preserve more detailed

spatial information about the plant and alleviate the above problems. These 3D data enable

quantitative assessment of 3D morphological traits of the plant (Li et al., 2022; Ni et al., 2021; Xi

et al., 2020), showing great potential for achieving high-precise segmentation of rapeseed plants.



However, the measurement error for the current widely used TOF and structured light sensors varies from 1 mm to cm-level due to the wiggling effect or the measured distance (Fürsattel et al., 2016; Fürsattel et al., 2017; Rauscher et al., 2016). The siliques have a slim shape with a much smaller width, requiring a high spatial resolution depiction. Thus, the 3D data of rapeseed plants acquired by TOF or structured light sensors may be less precise and unreliable. Handheld laser scanning (HLS) point cloud, one of the main forms of 3D data, is used to present detailed objects with high precision (Han et al., 2021), making it an ideal data source to improve the accuracy of silique phenotyping. Therefore, we proposed a fully annotated HLS rapeseed plant point clouds dataset in this study for high-precise 3D phenotyping.

It is very challenging to segment HLS rapeseed plant point clouds for two main reasons. First, HLS point clouds are highly dense, making it hard to assign semantic labels for each point in high spatial resolution. Second, silique distributions are disordered with strong scattering and mutual overlapping in a 3D space. A feasible segmentation pipeline is required to accurately detect and extract detailed traits of the rapeseed plants without compromising data resolution. Traditional methods investigate various 3D features defined by geometry properties and hand-crafted descriptions. They segment plant structures based on their 3D skeleton (Zermas et al., 2017), fast point feature histograms of the coordinate set (Sodhi et al., 2017; Wahabzada et al., 2015), or surface curvature and normals (Li et al., 2017; Li et al., 2013). However, these methods rely highly on predefined rules and prior knowledge of the segmented targets. Given that segmentation quality



is greatly influenced by the characteristics of different targets and parameter tuning (Vo et al., 2015), traditional methods are hence time and labor-consuming. In addition, the limited prior knowledge of plant morphology constrains traditional methods of plant 3D phenotyping on simple structures and traits (Gibbs et al., 2020; Paulus et al., 2014; Xiang et al., 2019).

On the other hand, deep-learning-based methods for point cloud segmentation have recently emerged to tackle this challenge. In contrast to leveraging prior knowledge, they learn features from input data in a data-driven manner. Benefitting from the advanced neural networks, deep learning methods outperform most traditional segmentation methods, showing great potential in plant 3D phenotyping (Guo et al., 2020a). Deep-learning-based methods on plant point cloud segmentation can be further divided into voxel-based and point-based methods (Guo et al., 2020b). Voxel-based methods transpose the point clouds into grids, and the regular voxel grids representation, like pixels in 2D images, can easily be analyzed by borrowing de-facto techniques from 2D counterparts, such as 3D convolution neural network (CNN) (Huang and You, 2016) and fully-convolutional point network (FCPN) (Rethage et al., 2018). Jin et al. (2020a) developed a voxel-based convolutional neural network (VCNN) to realize the classification and segmentation of maize at different growth stages. Though voxel-based methods can take in a dynamic number of input points, the traditional voxelization strategy, following the scheme of hard voxelization (HV) (Lang et al., 2019; Zhou et al., 2019; Zhou and Tuzel, 2018), may obscure the information of the raw inputs (Sec. 2.3). Besides, the paradigm of using convolution-based techniques on voxel grids



is hard to balance the performance and the computational cost. Point-based methods avoid obscuring invariances of the original data by directly learning features of each point using shared multi-layer perceptron (MLP) (Engelmann et al., 2019; Qi et al., 2017a; Yang et al., 2019), point-wise convolution (Hua et al., 2018; Thomas et al., 2019; Xu et al., 2021), or recurrent neural network (RNN) (Engelmann et al., 2017; Huang et al., 2018; Ye et al., 2018). Li et al. (2022) proposed a dual-function point-based network PlantNet to extend the semantic and instance segmentation of plant organs to three species. Jin et al. (2020b) developed a point-based fully convolutional neural network PFCN to resolve the difficulties of segmenting large-scale forest scenes. Turgut et al. (2022) evaluated how the synthetic plant data affected the performance of existing point-based deep learning algorithms. However, for most point-based methods, the network architecture and hyper-parameters are mainly designed for small-scale inputs due to the hardware limitations. Thus, the computation cost of point-based methods is highly sensitive to the number of input points, and a full-scale input will either increase the training speed or bring no performance improvements (Li et al., 2022). Down-sampling is necessary to reduce the number of points before being fed into a point-based network.

Overall, the recent efforts still remain infeasible facing our scenario: how to segment HLS rapeseed plants with tiny siliques while maintaining the complete spatial information? The main challenges lie in two aspects. First, the widely used hard voxelization in voxel-based methods and the down-sampling operation in point-based methods both result in great information loss to the



original data. Second, training and inferring on dense voxel grids or point clouds can bring an intolerable computational cost to the existing deep learning methods. Herein, we proposed a plant segmentation transformer (PST) aiming to segment HLS rapeseed plant point clouds in high spatial resolution while maintaining an acceptable inference speed. In summary, our main contributions are as follows.

(i) We propose a fully annotated HLS point clouds dataset containing 55 podding stage rapeseed plant samples. The proposed dataset has been manually supervised.

(ii) We proposed an end-to-end novel network plant segmentation transformer (PST) for dense HLS rapeseed plant point clouds semantic segmentation, which can segment tiny siliques in high-precise and low inference time. Specifically, we developed a dynamic voxel feature encoder (DVFE) to preserve and aggregate complete spatial information of the dense inputs. Further, to achieve a better trade-off between the segmentation performance and the inference time, we adopt the self-attention mechanism regarding its efficiency on point features learning and ability to capture neighbor contextual information of the data.

(iii) We integrated the proposed network with an optimized instance segmentation head and developed PST-PG to realize instance segmentation of HLS rapeseed plant point clouds.



## 2. Methods

### 2.1 HLS Rapeseed Plant Dataset

The proposed dataset was developed from eight rapeseed plant cultivars of ZD619, ZD622, ZD630, Sl512 CR3168, Bnw1.61/83, Hu135, and 8426016 at the podding stage. Specifically, the cultivation of ZD619, ZD622, and ZD630 was conducted at the Zijingang Campus of Zhejiang University, Hangzhou, Zhejiang Province, China. Sl512 CR3168, Bnw1.61/83, Hu135, and 8426016 were cultivated at the Changxing Agricultural Experiment Station of Zhejiang University, Hangzhou, Zhejiang Province, China. The eight cultivars chosen are the most common ones expected to turn into different structures at the podding stage. These structures can be further divided into three main types representing a wide range of various cultivars in terms of the growth of their branches and their first tiller location on the main stem. (Fig. 1) During the podding stage, the rapeseed plants were placed into plastic plots for data acquisition. Fifty-five rapeseed plant point clouds of eight cultivars at the podding stage were acquired through a handheld laser scanner (PRINCE775, SCANTECH (Hangzhou) Co., Hangzhou, China) with the maximum measurement error less than 1 mm. The rapeseed plant was scanned from multi-views until its spatial information was completely collected during the acquisition. The details of our dataset are given in Table 1.

Data labeling is labor-intensive but pivotal for deep learning networks training. In this study, we used CloudCompare (Girardeau-Montaut, 2015) to conduct the labeling process. Each point was annotated as the silique or non-silique class as the semantic labels by visual inspection. In



addition, points of different silique instances are also labeled by a unique number to enable the instance segmentation task. The labeling process of three representative samples is shown in Fig. 1. Different color represents different semantic classes, and each silique instance with a unique number was rendered with a random color.

**Table 1**

Description of the HLS rapeseed plant dataset. (The dataset is available upon request).

| Number of samples | Number of points | Plant coverage ([*lenth*, *width*, *height*] • cm) | Average silique proportion (%) | Average non-silique proportion (%) |
|---|---|---|---|---|
| 55 | $2.4×10^4$-$9.52×10^5$ | min: [11.08, 10.69, 53.48] max: [61.47, 69.97, 132.22] | 79.97 | 20.03 |

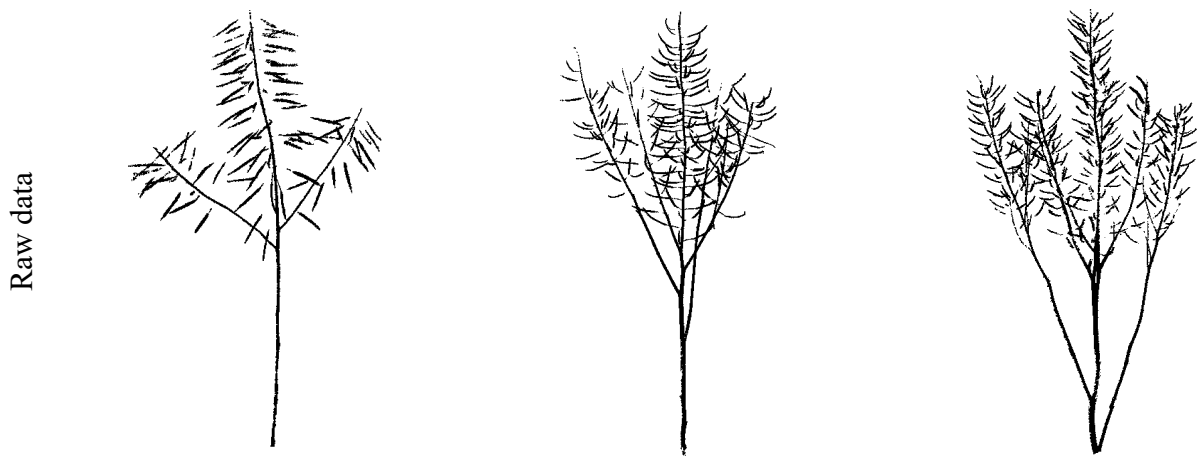

Raw data



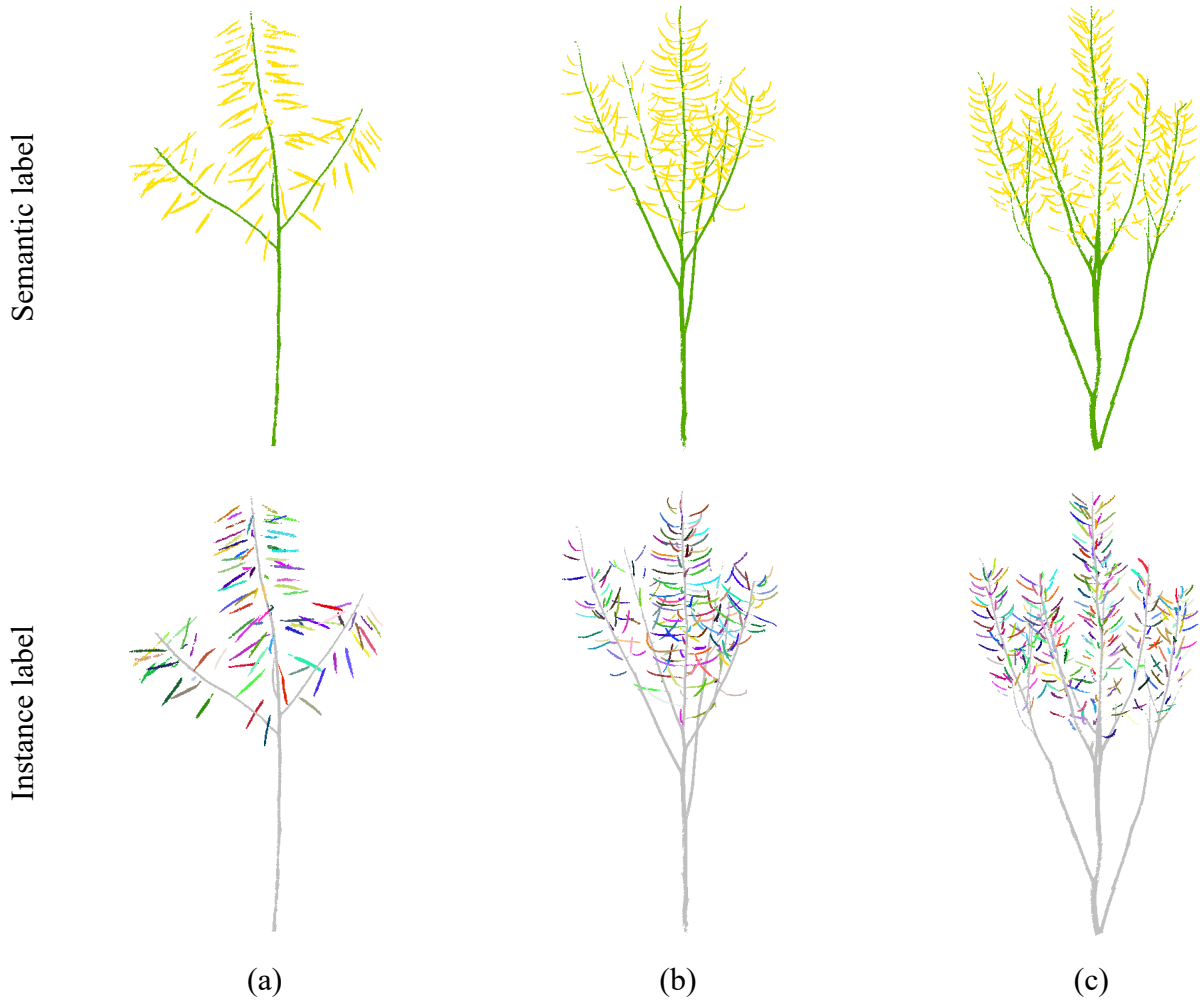

**Fig. 1.** Illustration of the data labeling of three samples representing three main structure types (S1, S2, and S3). S1 has a few branches, and all of them tiller from the main stem (a). S2 has multiple branches tillering from the main stem or other branches, and the first tiller location is in the middle of the main stem (b). S3 has more branches tillering from other branches than S2, and the first tiller location is at the bottom of the main stem (c). For semantic labeling, each point is annotated as the silique (yellow) or non-silique (green) class. For instance labeling, each silique instance is annotated with a unique number.



## 2.2 Network Overview

In order to segment the dense point clouds while maintaining the complete spatial information, the network should not only learn the features of every point in raw point clouds but also achieve an acceptable trade-off between the segmentation performance and the inference time. Thus, we use a dynamic voxel-based encoder to aggregate and primarily encode the per point feature of the raw inputs. In addition, we adopt the self-attention, a feature encoding mechanism that can extract the connections among features by computing attention in the data, based on the prior work (Fan et al., 2021; Liu et al., 2021) for further feature learning due to its efficiency of processing point clouds. Hence, we design our plant segmentation transformer (PST) to process dense HLS rapeseed plant point clouds. Fig. 2 overviews the network architecture of the PST, which follows an encoder-decoder pipeline and has three main components, i.e., the dynamic voxel feature encoder (DVFE), dual window sets attention, and dense feature recover decoder.

The input to the PST is a point set $\mathbb{P}$ with $N$ points. Each point set has 3D coordinates $P = \{p_i\} \in \mathbb{R}^{N \times 3}$, where $p_i = (x_i, y_i, z_i)$ and an initial feature set $F = \{f_i\} \in \mathbb{R}^{N \times C_0}$, where $C_0$ is the channel number of the input. The input is first fed into DVFE and embedded to the voxel set $\mathbb{V}$ with $N_V$ voxels. We denote $V = \{v_j\} \in \mathbb{Z}^{N_V \times 3}$ where $v_j = (x_j^V, y_j^V, z_j^V)$ and $F^V = \{f_j^V\} \in \mathbb{R}^{N_V \times C_1}$ as the voxel-wise coordinates and the high-dimensional feature set of the voxel $v_j$, respectively, where $C_1$ is the channel number of the voxels after being processed by DVFE.



After DVFE, the voxel set is partitioned into two sets of windows with a region shift mechanism (Fan et al., 2021). Each set contains multiple non-overlapping windows. We then apply multi-head self-attention to the voxel feature set $F^V = \{f_j^V\}$ in each window. The encoded voxel-wise output $G^V = \{g_j^V\} \in \mathbb{R}^{N_V \times C_2}$ is obtained after several dual window sets attention blocks, where $C_2$ is the channel number of the voxels processed during the attention blocks.

Lastly, $G^V = \{g_j^V\}$ is propagated into point-wise resolution and interacted with $F$ to form the final well encoded point feature set $G = \{g_i\} \in \mathbb{R}^{N \times (C_2 + C_0)}$. We then calculate the probability scores per point based on $G$ to get the final semantic label $S = \{s_i\}$ for each point. Furthermore, we integrate our PST with an instance segmentation head to form PST-PG (Sec. 2.6) and realize precise silique instance segmentation of rapeseed plants.

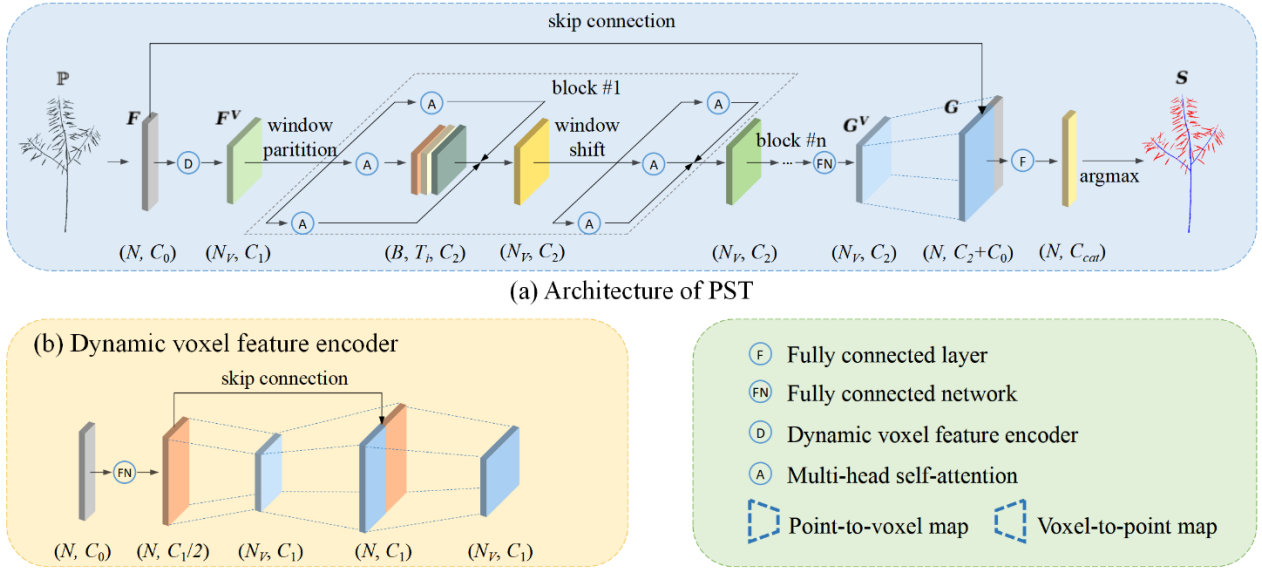

(a) Architecture of PST

(b) Dynamic voxel feature encoder

**Fig. 2.** Illustration of the PST architecture. (a) is the main component of PST, which follows an encoder-decoder pipeline. (b) is a demonstration of dynamic voxelization used in the encoding part.



## 2.3 Dynamic voxel feature encoder

Paradigm voxel-based methods usually voxelize the point clouds into dense grids and generate a one-to-many bi-directional map between every pair of $p_i$ and $v_j$. A voxel has a fixed capacity $Y$ of assigned points, so if more than $Y$ points are assigned to a voxel, they are sub-sampled to $Y$. Similarly, those less than $Y$ are zero-padded (Huang and You, 2016; Zhou and Tuzel, 2018). Such voxel representation, called hard voxelization (HV) (Zhou et al., 2019), naturally preserves the neighborhood context of 3D point clouds, leading to easy application of advanced techniques borrowed from 2D counterparts, ensuring a steady segmentation performance. However, its limitations are also intrinsic, i.e., (i) the stochastic dropout of points may cause information loss in voxelization, worsening discrimination of segmentation outcomes, especially for rapeseed plant point clouds with fine-grained traits; (ii) zero-padded voxels also occupy computation resources. To overcome the above issues, this study builds the dynamic voxel feature encoder (DVFE) based on dynamic voxelization (DV) (Zhou et al., 2019) for voxel feature embedding. $M_V(p_i)$ and $M_P(v_j)$ are defined as the mapping functions that assign each point $p_i$ to a voxel $v_j$ and gather the points within a voxel $v_j$ due to their 3D coordinates, respectively. For HV, when assigning more points than the voxel capacity $Y$, the points are sub-sampled, where extra points are dropped out. Therefore, for dropped out point $p_i$, the mapping function $M_V$ on it leads to an empty set ($\phi$). By contrast, instead of sub-sampling points to the fixed capacity $Y$ of a voxel, DV provides a complete map between $p_i$ and $v_j$ without information loss. Each point is assigned to a certain



voxel in terms of their spatial distance, and consequently, the number of points in the voxel is dynamic. To summarize, the difference between HV and DV is defined as follows (Zhou et al., 2019):

$$HV \begin{cases} M_V(p_i) = \begin{cases} \phi & \text{if } p_i \text{ is dropped out during sub-samping} \\ v_j & \text{if } p_i \text{ is preserved after sub-sampling} \end{cases} \\ M_P(v_j) = \{p_i | \, \forall p_i \in v_j\} \end{cases} \quad (1)$$

$$DV \begin{cases} M_V(p_i) = v_j, \forall i \\ M_P(v_j) = \{p_i | \, \forall p_i \in v_j\} \end{cases} \quad (2)$$

DVFE encodes the raw point-wise input to a voxel-wise embedding with a learned feature. Given an input point set $\mathbb{P}$, the 3D space is divided into voxel grids, and each point is assigned to the voxel it occupies. We denote $PC = \{pc_i\} \in \mathbb{R}^{N \times 3}$, where $pc_i = (x_i^C, y_i^C, z_i^C)$ is the $XYZ$ coordinates of the centroid of the points in the voxel $v_j$ that point $p_i$ belongs to, i.e.,

$$pc_i = \frac{1}{N^{v_j}} \sum_{k \in v_j} p_k \quad (3)$$

Where $N^{v_j}$ is the number of points in voxel $v_j$, and the mapping function in Eq. (3) is $M_P(v_j) = \{p_k\}$. With $PC$ and $V$ (Sec. 2.2), the raw input feature $F$ is now augmented as $\hat{F} = \{\hat{f}_i\}$, where $\hat{f}_i = (x_i, y_i, z_i, x_i - x_i^C, y_i - y_i^C, z_i - z_i^C, x_i - x_j^V, y_i - y_j^V, z_i - z_j^V) \in \mathbb{R}^9$. Next, two consecutive VFE layers (Sindagi et al., 2019; Zhou and Tuzel, 2018), each consisting of a fully connected network, are applied to obtain a voxel-wise high-dimensional feature set $F^V = \{f_j^V\}$, the above operation can be defined as:

$$\hat{F}^V = \Lambda_{M_V}(FCN(\hat{F})) \quad (4)$$

$$F^V = \Lambda_{M_V}(FCN(Concat(V_{M_P}(\hat{F}^V), FCN(\hat{F})))) \quad (5)$$



Eq. (4) and Eq. (5) are the first and the second VFE layer, respectively, where $\Lambda_{M_V}$ refers to the aggregation function, i.e., average, max, or sum in terms of the mapping function $M_V$. $\vee_{M_P}$ refers to the propagation function that recovers voxel-wise feature set to point-wise feature set in terms of $M_P$. $FCN$ represents the fully connected network composed of a fully connected layer, a batch normalization layer, and an activation function. The first VFE layer aggregates the decorated point features belonging to a specific voxel using the max-pooling (i.e., $\Lambda_{M_V}$) to a voxel-wise feature set $\hat{F}^V$. The second VFE layer propagates $\hat{F}^V$ using $\vee_{M_P}$ and concatenates it with the learned point features. Finally, $\Lambda_{M_V}$ is used again to aggregate the final feature embedding of DVFE. Simplified VFE layers is shown in Fig. 3(a). Note that a sequent operation of $\vee_{M_P}(\Lambda_{M_V}(F)$ is not equal to $F$ (Fig. 3(b)).

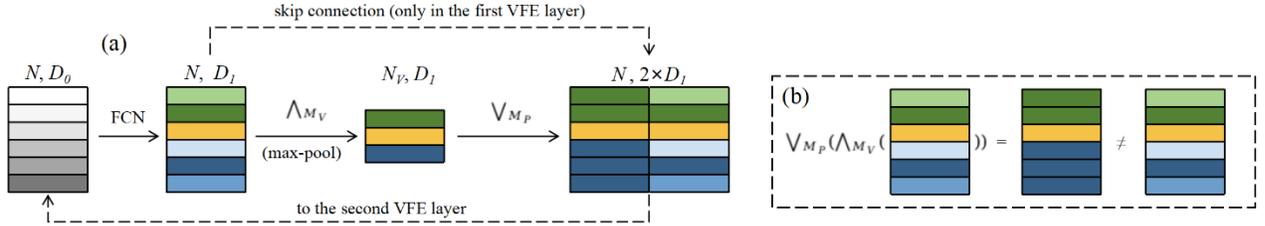

**Fig. 3.** Illustration of the simplified VFE layers in DVFE.

For simplicity and clarity, we adopt the coordinates $P$ as the initial features $F$ of the input fed into DVFE (i.e., $P=F$) in this study. Other features like reflectance, RGB, or normal can easily be used by changing the channel numbers.

## 2.4 Dual window sets attention

After the DVFE, the voxel features have captured a preliminary abstraction of the aggregated



point features; however, it lacks the contextual information among the voxels. As discussed in Sec. 2.2, we adopt the self-attention mechanism to compute attention among the voxels to capture the neighbor context for voxel feature learning. Directly applying the self-attention on a global scale of voxels is undesirable, leading to a dilemma between a precise performance and a scalable computation cost since immense representatives of voxels are required for satisfying results. Therefore, we adopt the shifted-window self-attention following the prior work in Single Stride Transformer (Fan et al., 2021), which has been proven to be very effective in the 2D image and autonomous driving scenarios.

### 2.4.1 Window partitioning and shifting

PST partitions the voxel grids into non-overlapping axis-aligned windows for computing self-attention locally in each window. The number of voxels divided into the window is controlled by hyper-parameters. Supposing the size of the window is $(l_x^W, l_y^W, l_z^W)$. Thus each window contains $l_x^W \times l_y^W \times l_z^W$ voxels for the calculation of their semantic affinities. Further, the origin partitioned window set is shifted at a Euclidean distance of $(\frac{l_x^W}{2}, \frac{l_y^W}{2}, \frac{l_z^W}{2})$ to form a new set for building connections across windows and enriching the contextual information. Consequently, two sets containing multiple windows, illustrated as "dual window sets", are obtained.

### 2.4.2 Self-attention in dual window sets

As the points are distributed sparsely in the 3D space, only the voxels assigned at least one



point are seen as valid, while the unoccupied ones are invalid. Thus the number of valid voxels in each window varies. To handle such sparsity, PST divides the windows into sub-batches regarding the number of valid voxels within. Given the number of total voxels in a window $N_V^W = l_x^W \times l_y^W \times l_z^W$, the sub-batches are divided as in Table 2.:

**Table 2**

Sub-batches in training and inference phases

| Phase | Sub-batch | Number of valid voxels | Voxels padding |
|-------|-----------|------------------------|----------------|
| training | 1 | $0\sim0.25N_V^W$ | $0.25N_V^W$ |
| | 2 | $0.25\sim0.5N_V^W$ | $0.5N_V^W$ |
| | 3 | $0.5\sim1N_V^W$ | $0.9N_V^W$ |
| inference | 1 | $0\sim0.25N_V^W$ | $0.25N_V^W$ |
| | 2 | $0.25\sim0.5N_V^W$ | $0.5N_V^W$ |
| | 3 | $0.5\sim0.9N_V^W$ | $0.9N_V^W$ |
| | 4 | $0.9\sim1N_V^W$ | $N_V^W$ |

The windows containing a similar level of valid voxels are divided into the same sub-batch. Then the number of voxels in each window is padded to the same value (Table 2. voxel padding) so that the self-attention within each window in a sub-batch can be calculated in parallel. Specifically, to add variance in the training phase, we set the highest padding level as $0.9N_V^W$, which means a window with more than $0.9N_V^W$ valid voxels will be randomly sampled to $0.9N_V^W$. Finally, the self-attention in dual window sets are computed as:

$$\text{set 1} \begin{cases} \widetilde{F}_i^V = MSA(LN(F_{i-1}^V), PE(V_{i-1})) + F_{i-1}^V \\ F_i^V = MLP\left(\widetilde{F}_i^V\right) + \widetilde{F}_i^V \end{cases} \tag{6}$$



$$\text{set 2} \begin{cases} \widetilde{F}_{i+1}^V = MSA(LN(F_i^V), PE(V_i)) + F_i^V \\ F_{i+1}^V = MLP(\widetilde{F}_{i+1}^V) + \widetilde{F}_{i+1}^V \end{cases} \tag{7}$$

where $MSA$ is multi-head self-attention module, $LN$ is layer normalization, $PE$ denotes the position encoding function in (Carion et al., 2020). $\widetilde{F}_i^V$ and $F_i^V$ are the voxel-wise output feature map of the $MSA$ and $MLP$ module in block $i$ (in this study $i$=6), respectively.

## 2.5 Dense feature propagation

The final output $G^V$ is well encoded after several dual window sets attention blocks. To obtain a dense point-wise encoded feature set $G$ for computing the semantic labels $S$ per point, we first recover $G^V$ to its point-wise resolution using the propagation function $\bigvee_{M_P}$ as in DVFE (Sec. 2.3). Second, we concatenate the propagated feature map with a learned input $F$ to build up interaction between them and enrich the semantic granularity of $G$. Then, $G$ with the dimension of $N \times C_2$ is transformed to $N \times C_{cls}$ by a fully connected layer, where $C_{cls}$ is the number of semantic classes. In this study we set $C_{cls}$=2 (i.e., siliques and non-siliques). Finally, the probability scores per point for all classes are computed by carrying out a softmax operation, and the class with the highest probability is assigned to that point. The above operation can be defined as:

$$G = Concat(\bigvee_{M_P}(G^V), MLP(F)) \tag{8}$$

$$S = Argmax(Softmax(FC(G))) \tag{9}$$

In the training phase, the network takes random patches from the training dataset at an amount



of batch size in each iteration. Consequently, the training patches may not cover all the points in an input point cloud. Such training approaches ensure the robustness of the network. However, in the inference phase, to obtain complete segmentation labels of the input point cloud without losing a point, we adopt a region-slide strategy with an overlap to traverse every point. The final semantic label assigned to each point is obtained on the averaged probability scores.

## 2.6 Integration with instance segmentation

A two-stage pipeline considering instance segmentation as a subsequent clustering stage after semantic segmentation is widely used in the design of an end-to-end instance segmentation network (Elich et al., 2019; Han et al., 2020; Mo et al., 2019; Pham et al., 2019; Wang et al., 2018). In such a pipeline, the points with semantic labels predicted in the first stage are grouped into instances by an instance segmentation head in the second stage. Inspired by Jiang et al. (2020), we combine PST with the instance segmentation head in PointGroup (PG) and form PST-PG (Fig. 4) to achieve the instance segmentation of siliques in rapeseed plants. The contextual and morphology traits are well abstracted by PST, which provides discriminative point-wise features for the subsequent processing in the second stage.

We choose the instance segmentation head in PG as the base network of our second stage mainly for two reasons: (i) PG is developed in a hybrid-model manner, making it flexible when hybridizing with other models (i.e., PST) served in the first or second stage; (ii) PG leverages the void space between instances to increase the performance of instance segmentation. Since the void



space between each silique in a rapeseed plant at the podding stage is also quite regular, it can be an effective backup in our scenario.

### 2.6.1 Instance segmentation head in PointGroup

This section briefly revisits the instance segmentation head in PG for completeness. The network mainly contains three parts: (i) learn a per point offset vector to shift each point to its corresponding instance centroid; (ii) use a clustering algorithm to group points with semantic predictions into candidate clusters in the original coordinate space and shifted coordinate space; (iii) predict the scores for each candidate using ScoreNet to select the proper cluster.

Given a point $i$, the void space-based clustering algorithm neighbors the points within an $r$-sphere centered at $p_i = (x_i, y_i, z_i)$, where $r$ serves as a spatial threshold, and groups points with the same semantic labels as $i$ into the same candidate cluster. Here, points with distances larger than $r$ or in different classes will not be grouped. However, clustering only in the original coordinate space may wrongly group the same-class instances close to each other. Thus, a sub-branch network is trained to learn an offset $OS = \{os_i\} \in \mathbb{R}^{N \times 3}$ for shifting each point in $P = \{p_i\}$ towards its instance centroid. In the shifted coordinate set $P^S = P + OS \in \mathbb{R}^{N \times 3}$, the void space between the instances increases so that the adjacent same-class instances can be discriminated better.

Denote $C^O$ and $C^S$ as the clustering results on the original coordinate set $P$ and the shifted coordinate set $P^S$, respectively. PointGroup then constructs a ScoreNet to predict a score for each candidate cluster in $C = C^O \cup C^S$ to describe their quality. In the inference phase, the non-



maximum suppression (NMS) is adopted on the clusters to quantify their quality so that the acceptable ones can be added to the final instance predictions.

As for the loss functions, we use a standard cross-entropy loss $L_{c\_sem}$ for the semantic branch (i.e., PST) in the first stage. In the second stage, we adopt the same settings as in PG. Specifically, for the offset prediction branch, two loss functions are adopted. One is a $L_1$ regression loss $L_{o\_reg}$ to constrain the $L_1$ norm between each point and its corresponding instance centroid. The other is a direction loss $L_{o\_dir}$ (Lahoud et al., 2019) to ensure each point moves towards its centroid. For ScoreNet, the loss function is a binary cross-entropy loss as $L_{c\_score}$.

### 2.6.2 Variants of PST-PG

**V-PST-PG**

We build the base version of PST-PG, called Vanilla (V)-PST-PG, as in Fig. 4. In the implementation of V-PST-PG, we feed $F$ in two branches, one for semantic segmentation (i.e., PST) to obtain class labels $S$, the other for predicting the offset $OS$ to shift the original coordinates $P$ to $P^S$. As we are only concerned with the silique instances, the non-silique predictions from PST are masked and have no effect during the clustering part.

After clustering, supposing the total number of candidate clusters is $M$ and $N_i^C$ denotes the number of points in cluster $C_i$, that is $C = \{C_i\} \in \mathbb{R}^{M \times N_i^C \times 3}$. We then gather the high-dimensional point feature $G = \{g_i\}$ from PST, followed by an extra $MLP$ layer to form a cluster-wise feature set $F^C = \{f_i^C\} \in \mathbb{R}^{M \times N_i^C \times C_3}$ as the initial feature of $C$ for ScoreNet, where $C_3$ is the channel



number of $G$ after the $MLP$ layer. The final cluster scores $S^C = \{s_i^C\} \in \mathbb{R}^M$ are obtained as:

$$S^C = Sigmoid\left(MLP\left(ScoreNet(F^C, C)\right)\right) \tag{10}$$

V-PST-PG is trained in an end-to end manner following the initial configuration as in PG. The clustering and ScoreNet are not activated in training until the semantic and the offset prediction branches arrive at the preparation epoch. After the preparation epoch, all components in V-PST-PG are trained simultaneously. The total loss function for V-PST-PG is defined as:

$$L_{v\_ins} = \begin{cases} L_{c\_sem} + L_{o\_reg} + L_{o\_dir} & \text{epoch} \leq \text{preparation epoch} \\ L_{c\_sem} + L_{o\_reg} + L_{o\_dir} + L_{c\_score} & \text{epoch} > \text{preparation epoch} \end{cases} \tag{11}$$

**F-PST-PG**

Inspired by (Brock et al., 2017), as long as the semantic branch is capable of providing ideal and stable semantic predictions for the subsequent training, it is unnecessary to have PST participate after the preparation epoch. Thus, we freeze the PST by excluding them from the backward pass after the preparation epoch to form F-PST-PG (i.e., PG with frozen PST). Unlike V-PST-PG, where the semantic output $S$ changes after every training iteration, in the implementation of F-PST-PG, once PST is well trained, the semantic output $S$ is fixed and serves as the fixed-learned supervision for the instance segmentation head. The total loss function for F-PST-PG is defined as:

$$L_{f\_ins} = \begin{cases} L_{c\_sem} + L_{o\_reg} + L_{o\_dir} & \text{epoch} \leq \text{preparation epoch} \\ L_{o\_reg} + L_{o\_dir} + L_{c\_score} & \text{epoch} > \text{preparation epoch} \end{cases} \tag{12}$$



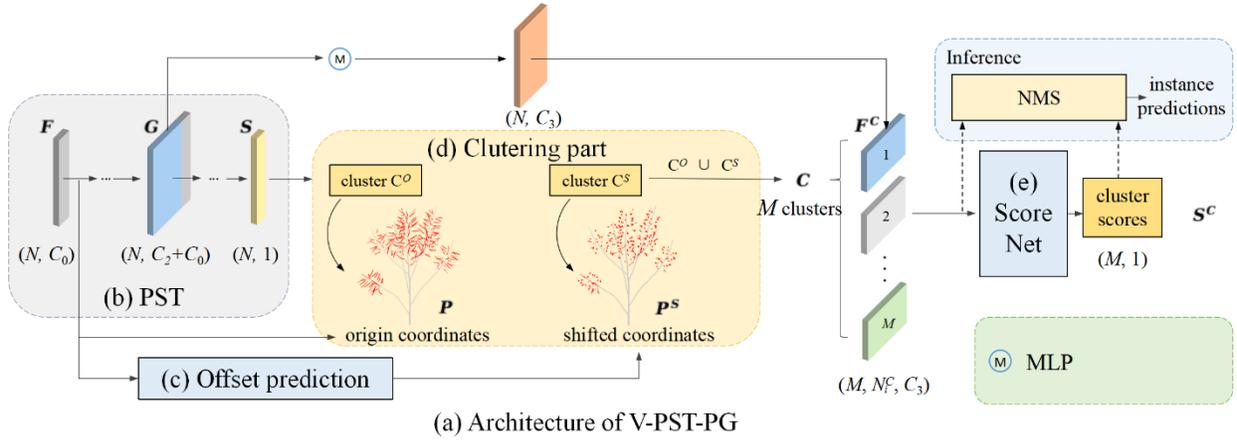

(a) Architecture of V-PST-PG

**Fig. 4.** Illustration of V-PST-PT (a), which consists of four parts: PST (b), offset prediction (c), the clustering part (d), and ScoreNet (e).

# 3 Experiment

## 3.1 Dataset preparation

The HLS rapeseed plant dataset is split into training, validation, and testing sets. Specifically, sample 1-40 are used for training, sample 41-49 are used for validation, and sample 50-55 are used for testing. In the ablation study and hyper-parameters choosing, we train on the training set and report results on the validation set. In the evaluation and comparison with other approaches, we train on the training set and report the results on the testing set. We also implement six-fold cross-validation on sample 1-55.

In the training phase, we use a fixed-size cubic to crop the whole plant point clouds into patches, each treated independently in the network. And in the validation and testing phases, we also partition each whole plant point cloud into patches and use a region-slide inference strategy to



ensure that every point is predicted by the network.

To enrich the training data, we partition the point clouds with two offset values (0 and 8 cm), resulting in two sets of different patches. Therefore, the actual annotated data for training, validating, and testing extracted from the integrated plant are enriched at a high level, assuring a large amount of data for the deep learning algorithm. The choice of patch size depends on the properties of the input data. To ensure each patch encompasses moderate semantic information, we set the patch length to 16cm during the study.

## 3.2 Network training and testing

All the experiments in this study are conducted on the Pytorch platform. Specifically, to achieve a fair comparison, we design the proposed network PST/PST-PG and compare different baseline models based on the open framework MMDetection3D. For the network setting and parameters selection of the baseline models, we follow the configuration from their original papers.

We train PST using AdamW optimizer with a weight decay of 0.05 and the cyclic learning rate schedule. The base learning rate is $10^{-5}$, and the maximum learning rate is $10^{-3}$. During the training, the batch size is set to 4. The network is evaluated every two epochs on the validation set and trained until the loss is stable both on the training and validation sets. For testing, the batch size is set to 1, and the learned parameters of PST for testing are determined in terms of the lowest loss on the validation set.



### 3.3 Implementation details

As for the implementation details of PST, in dynamic voxel feature encoder (DVFE), the voxel size used in dynamic voxelization (DV) is 0.6cm×0.6cm×0.25cm, and the aggregate function is max-pooling. In dual window sets attention, the window size is 6×6×12, which means the total number of voxels within the window (i.e. $N_V^W$) is 432. In dense feature propagation, the overlap for region-slide inference is 8cm (half of the patch size). Thus a validation or testing patch is predicted twice to obtain the final average probability scores.

For instance segmentation, we build up the instance segmentation head and implement the second stage of PST-PG adopting the same settings in PointGroup. Please refer to (Jiang et al., 2020) for details.

### 3.4 Evaluation metrics

In this study, we evaluate the semantic segmentation network (i.e., PST) and the instance segmentation network (i.e., PST-PG) separately.

For semantic segmentation, we evaluate Intersection-over-Union (IoU), Precision (Prec), Recall (Rec), and F1-score four class-level metrics and one global metric, overall accuracy (oAcc). Specifically, for each semantic class, IoU is known for measuring overlap between predicted points and ground truth points. Prec and Rec measure the correctly predicted points to the total predicted points and total ground truth points, respectively. F1-score is defined as the harmonic mean of Prec



and Rec. Across all the classes, oAcc is the proportion of total correctly predicted points to the total number of points. We also include these results for comparison (Sec. 4.1). The five metrics are defined as:

$$IOU = \frac{TP_C}{TP_C + FP_C + FN_C}$$

(13)

$$Prec = \frac{TP_C}{TP_C + FP_C}$$

(14)

$$Rec = \frac{TP_C}{TP_C + FN_C}$$

(15)

$$F1 = 2\frac{Prec \cdot Rec}{Prec + Rec}$$

(16)

$$oAcc = \frac{TP + TN}{TP + TN + FP + FN}$$

(17)

where $TP_C$, $FP_C$, $FN_C$ are the number of true positive, false positive, and false negative points for a certain classs $Cls$, respectively. In this study, $Cls \in \{$silique, non-silique$\}$.

For instance segmentation, we use mean precision (mPrec), mean recall (mRec), mean coverage (mCov) and mean weighted coverage (mWCov) (Li et al., 2022; Liu et al., 2017; Ren and Zemel, 2017; Wang et al., 2019a; Zhuo et al., 2017). Specifically, mPrec$_\theta$, and mRec$_\theta$ denote the mPrec and mRec with the IoU threshold set to $\theta$. mCov is the average IoU of instance prediction matched with ground truth. mWCov is calculated as mCov weighted by the size of each ground truth instance. The four metrics are defined as:

$$mPrec_\theta = \frac{TP_\theta^{ins}}{|O|}$$



$$(18)$$

$$mRec_\theta = \frac{TP_\theta^{ins}}{|R|}$$

$$(19)$$

$$mCov = \frac{1}{|R|} \sum_{i=1}^{|R|} \max_j IoU(P_i^R, P_j^O)$$

$$(20)$$

$$mWCov = \sum_{i=1}^{|R|} w_i \max_j IoU(P_i^R, P_j^O)$$

$$(21)$$

$$w_i = \frac{|P_i^R|}{\sum_k |P_k^R|}$$

$$(22)$$

where $TP_\theta^{ins}$ is the number of predicted instance having an IoU larger than $\theta$ with the ground truth. $|R|$ and $|O|$ is the number of all instances in the ground truth and prediction, respectively. In Eq. 19 and Eq.20, $|P_i^R|$ is the number of points in the $i$-th ground truth instance, and $|P_j^O|$ is the number of points in the $j$-th predicted instance.

## 4. Results

### 4.1 Semantic segmentation

PST outperforms all the counterpart networks with the highest performance in mean IoU (93.96%), mean precision (97.29%), mean recall (96.52%), mean F1-score (96.88%), and overall accuracy (97.07%) (Table 3). The improvement of these metrics compared to the second-best results achieved by PAConv are 7.62, 3.28, 4.8, 4.25, and 3.88 percentage points. Fig. 5 illustrates



the performance of four networks on the testing set. Although siliques are of small size and scattered closely over the rapeseed plants' branches (i.e., non-silique) in the 3D space, making them hard to be extracted, PST still achieves the best sensitivity and accuracy in distinguishing them. PointNet++ (MSG) (Qi et al., 2017b) serves as the most effective network among the PointNet family, often failing to recognize the branches in the canopy and consider all the objects as an ensemble in that region, achieving the worst results across all the networks. PAConv (Xu et al., 2021) is built up based on PointNet using an adaptive convolution mechanism to learn contextual information. It has a better discernibility in the canopy than the other two counterparts. However, we notice that in the junction region where the stem tillers, the performance of PAConv deteriorates as the structure becomes complex. The graph-based method DGCNN considers point clouds as spatial graphs and focuses on the edge information of the constructed graphs. The performance of DGCNN (Wang et al., 2019b) is placed between PointNet++ and PAConv as it may ignore the branches and often confuse the intra-class points.

**Table 3**

The comparison of semantic segmentation across the four networks. The best results are in boldface.

| Method | | IoU (%) | Prec (%) | Rec (%) | F1 (%) | oAcc (%) |
|---|---|---|---|---|---|---|
| PointNet++ | silique | 85.33 | 89.20 | 95.16 | 92.08 | |
| | non-silique | 75.71 | 91.34 | 81.56 | 86.18 | |
| | mean | 80.52 | 90.27 | 88.36 | 89.13 | 89.93 |
| PAConv | silique | 89.86 | 91.46 | 98.09 | 94.66 | |



| Method | | | | | | |
|---|---|---|---|---|---|---|
| | non-silique | 82.82 | 96.55 | 85.35 | 90.60 | |
| | mean | 86.34 | 94.01 | 91.72 | 92.63 | 93.19 |
| DGCNN | silique | 86.71 | 89.77 | 96.22 | 92.88 | |
| | non-silique | 77.76 | 93.17 | 82.46 | 87.49 | |
| | mean | 82.24 | 91.47 | 89.34 | 90.19 | 90.93 |
| PST (Ours) | silique | 95.40 | 96.43 | 98.89 | 97.65 | |
| | non-silique | 92.51 | 98.15 | 94.15 | 96.11 | |
| | mean | **93.96** | **97.29** | **96.52** | **96.88** | **97.07** |

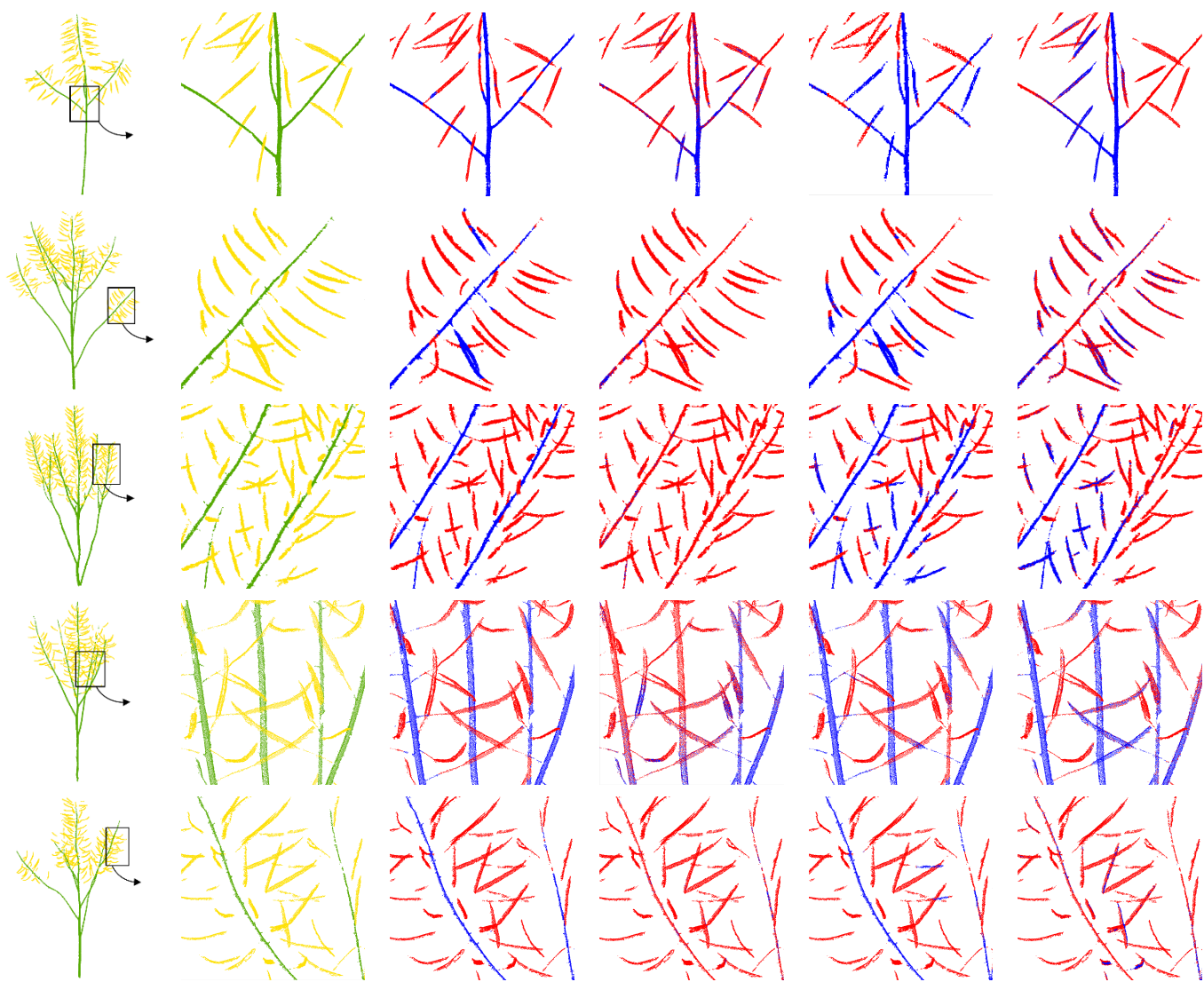



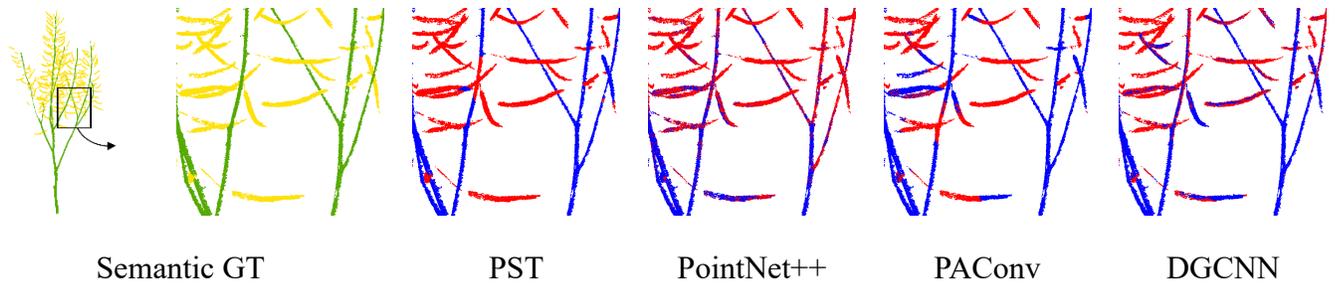

| Semantic GT | PST | PointNet++ | PAConv | DGCNN |

**Fig. 5.** Illustration of the qualitative results on the testing set by PST, PointNet++, PAConv, and DGCNN.

**Six-fold cross validation**

We evaluate PST by the six-fold cross-validation on the whole dataset (Fig. 6). Each sample is treated as the testing data once. PST performs more stable on segmenting siliques than non-siliques slightly with a lower standard deviation (STEDV) of IoU (1.46) and F1-score (0.79). In summary, PST achieves satisfying stability on HLS rapeseed plant data with 1.23 and 0.68 on STEDV of mean IoU (mIoU) and overall accuracy (oAcc), respectively.



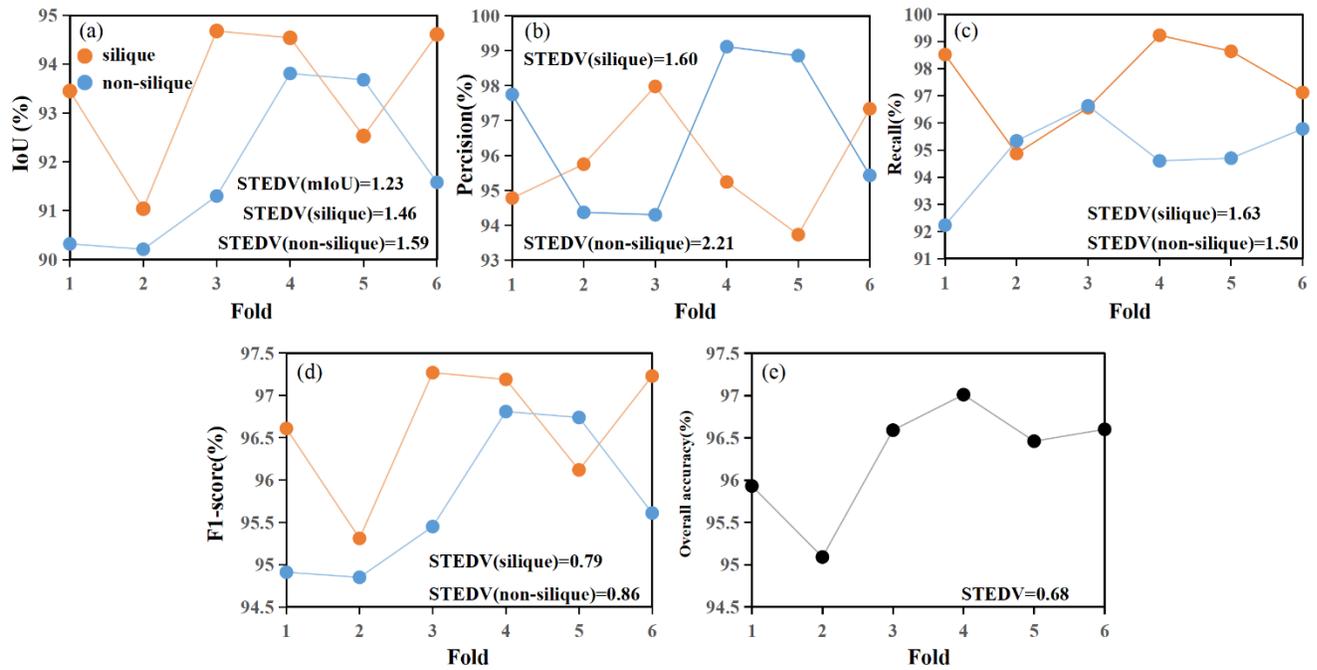

**Fig. 6.** Results of IoU (a), precision (b), recall (c), F1-score (d) and overall accuracy (e) under the six-fold cross validation.

## 4.2 Inference time analysis

For the potential of being applied in phenotyping, we are concerned with the inference time of a deep learning network. We test each method several times on an NVIDIA GeForce RTX 3090 GPU to get the average inference time listed in Table 4. PST takes the full scale of point clouds as inputs without adopting down-sampling strategies, while the other counterparts only accept fix size inputs for inference, which may consume computation resources on extra traversing in order to obtain the same resolution outputs as PST. PST ranks front among different testing groups with 156.2ms to predict a full-scale point cloud, reflecting a high potential for being used in real-time



phenotyping tasks.

**Table 4**

The inference time per point cloud of the four networks. We sample the points in a patch to a certain number (i.e., Input points) to satisfy the input requirement of that network. Full-scale means we use the original point clouds without sampling as inputs.

| Method | Input points | mIoU (%) | Inference time per point cloud (ms) |
|--------|-------------|----------|-------------------------------------|
| PointNet++ | 8k | 64.84 | 58.5 |
| PointNet++ | 38.4k | 80.52 | 91.9 |
| PAConv | 8k | 85.59 | 195.7 |
| PAConv | 20.4k | 86.34 | 139.3 |
| DGCNN | 8k | 80.37 | 258.4 |
| DGCNN | 20.4k | 82.24 | 336.7 |
| DGCNN | 38.4k | 77.71 | 497.5 |
| PST(Ours) | full-scale | 93.96 | 156.2 |

**4.3 Instance segmentation**

To evaluate the performance of PST as being integrated with the current instance segmentation head, we compare the instance segmentation results of the original PointGroup (PG), vanilla PST-PG, and frozen PST-PG listed in Table 5. The performance of two revised PST-PG networks has both improved compared to the original PG in mCov, mWCov, and mPrec, mRec with higher IoU threshold, demonstrating the effectiveness of PST for passing discriminative point features to the subsequent network. Specifically, F-PST-PG reaches 89.51% on mCov, 89.85% on mWCov, 88.83%



on mPrec$_{90}$ and 82.53% on mRec$_{90}$. The visual illustrations of F-PST-PG over three representative

samples in the testing set are shown in Fig. 7.

**Table 5**

The comparison of instance segmentation of the original PG, V-PST-PG, and F-PST-PG. The best

results are in boldface.

| Method | mCov | mWCov | mPrec$_{50}$ | mRec$_{50}$ | mPrec$_{75}$ | mRec$_{75}$ | mPrec$_{90}$ | mRec$_{90}$ |
|---|---|---|---|---|---|---|---|---|
| PG | 86.58 | 87.64 | **97.41** | 84.43 | 90.72 | 78.63 | 86.84 | 76.63 |
| V-PST-PG | 89.29 | 89.66 | 97.10 | 89.79 | **91.30** | 84.42 | 88.21 | 81.57 |
| F-PST-PG | **89.51** | **89.85** | 96.66 | **90.05** | 91.27 | **85.03** | **88.83** | **82.53** |

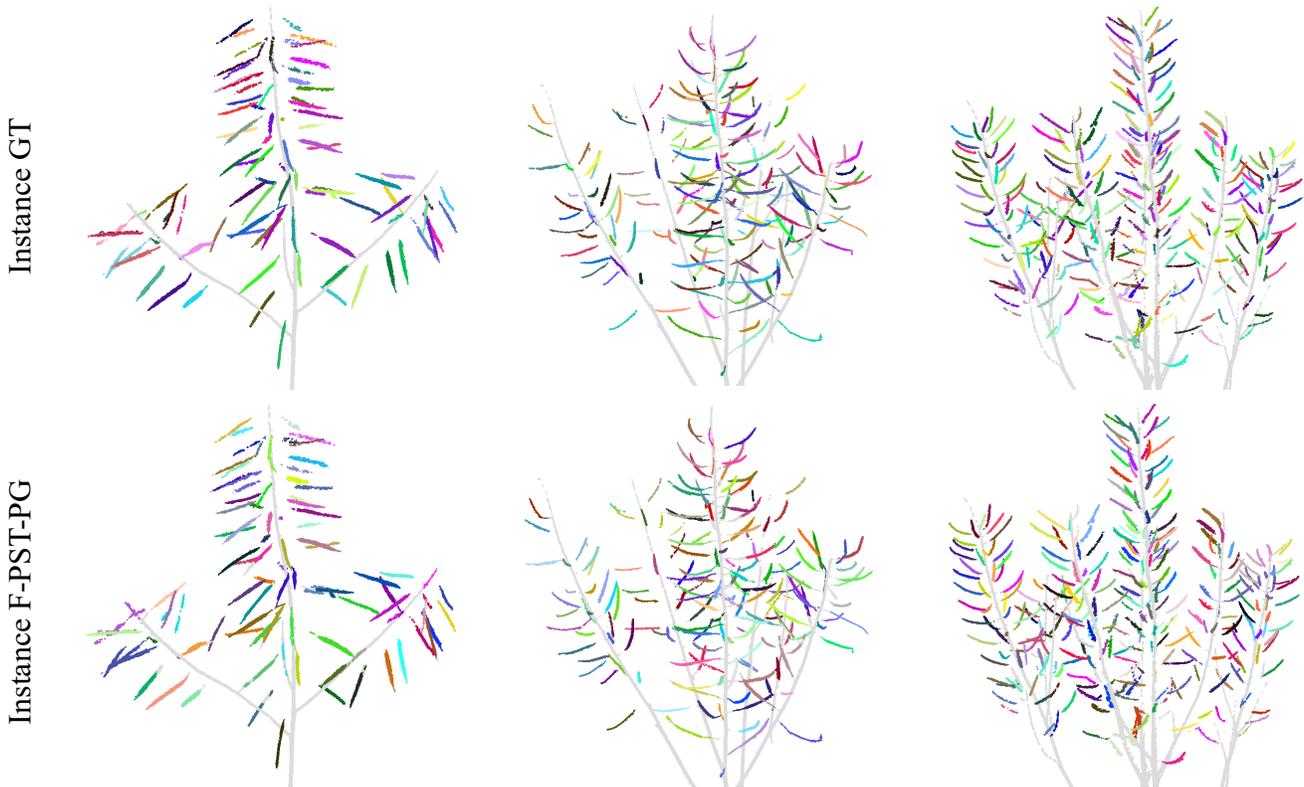



**Fig. 7.** Illustration of the qualitative results on three representative samples in the testing set by F-PST-PG. As we are only concerned with silique instances of rapeseed plants, each silique instance is labeled with a random color, and the non-silique parts are ignored and set as gray.

We further count the number of silique instances detected by the original PG and F-PST-PG with the ground truth under the first two strict standards (Fig. 8). A predicted silique instance is counted when it has an IoU larger than 75% (Fig. 8(a)) or 90% (Fig. 8(b)) with the ground truth instance collections. For both situations, F-PST-PG performs better than the original PG. The RMSE of F-PST-PG is 21.09 when IoU>75% and 25.87 when IoU>90% on the testing samples with the average silique instance number 109, which means the undetected silique instance should be less than 21.09 (when IoU>75%) or 25.87 (when IoU>90%) when the number of silique instances of rapeseed plants is 109. We also notice that when detecting samples with more than 200 siliques, F-PST-PG outperforms PG by a big margin, indicating the effectiveness of F-PST-PG when facing complex samples.

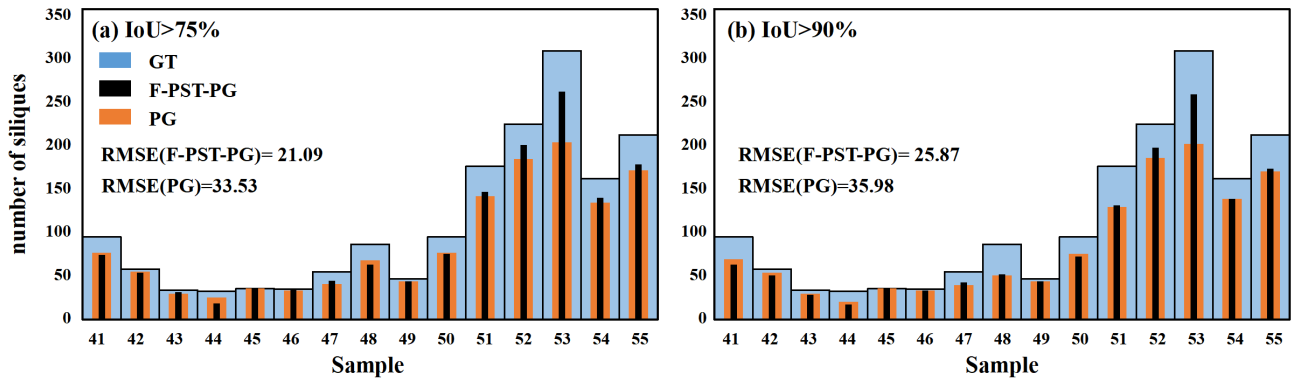

**Fig. 8.** The number of silique instances detected by F-PST-PG on the validation and testing samples



compared with the original PG and ground truth. The blue bar (widest) is the ground truth instance number for each sample. The orange bar is the instance number detected by the original PG. The black bar (narrowest) is the instance number detected by F-PST-PG.

## 5 Discussion

### 5.1 Ablation study on stacked point features in DVFE

As we discussed in Sec. 2.3, the raw point feature set is augmented by stacking each point feature with extra information before being fed into DVFE. Here, we conduct an ablation study on the validation set to analyze the choice of augmented features in DVFE. The features provided by cluster centroid per point are chosen as primitives. Besides, we consider using the corresponding voxel centroid (i.e., voxel coordinate) and L2 norm of each point for feature augmentation.

Table 6 reports the performance of PST under different settings. This indicates that using the combination of cluster and voxel centroid ensures the best performance of PST, where mean IoU on the validation set deteriorates when adding L2 norm per point. Indeed, L2 norms are calculated based on the point coordinates on a global scale, while the adoption of two centroids tends to capture the contextual information within a local region (i.e., a cluster of points and a voxel). Consequently, the local feature values provided by these two centroids are relatively small compared to L2 norms, resulting in an insufficient output with imbalance feature weights.

**Table 6**



Ablation results for PST with different augmented features on the validation set. $x^C, y^C, z^C$ refers to the cluster centroid. $x^V, y^V, z^V$ refers to the voxel centroid. $\parallel p \parallel_2$ refers to L2 norm.

| $x^C, y^C, z^C$ | $x^V, y^V, z^V$ | $\parallel p \parallel_2$ | mIoU |
|:---:|:---:|:---:|:---:|
| √ | | | 94.00 |
| √ | | √ | 93.85 |
| √ | √ | | **94.65** |
| √ | √ | √ | 93.06 |

**5.2 The voxel size in dynamic voxelization**

The variations in voxel size used in dynamic voxelization can cause different semantic distributions between the original point cloud and its counterpart after voxelization. In this study, each voxel is represented as $l \times w \times h$ (*length × width × height*). We compare the semantic proportion of silique and non-silique after voxelization under four different situations: $l=w>h$, $l=h>$w, $l<w=h$ and $l=w=h$. Specifically, all the point clouds are normalized in a 1m³ cubic during the experiment. Therefore, we use the voxel with 0.6cm, 0.6cm, 0.25 cm for $l=w>h$, $l=h>w$, $l<w=h$ and the voxel with 0.45cm × 0.45cm × 0.45 cm for $l=w=h$ since they output a similar number of voxels in a 1m³ cubic, leading to a fair comparison of four situations with similar voxel-wise resolution.

As shown in Fig. 9, the proportion of both silique and non-silique after voxelization under the voxel with $l=w>h$ achieves the best match with their initial proportion, which means using a flat voxel outperforms others, and even a normalized voxel when fitting the distribution of the original



data. In addition, the bottle-up growth pattern of rapeseed plants makes them more distinguishable in terms of a height-aligned perspective.

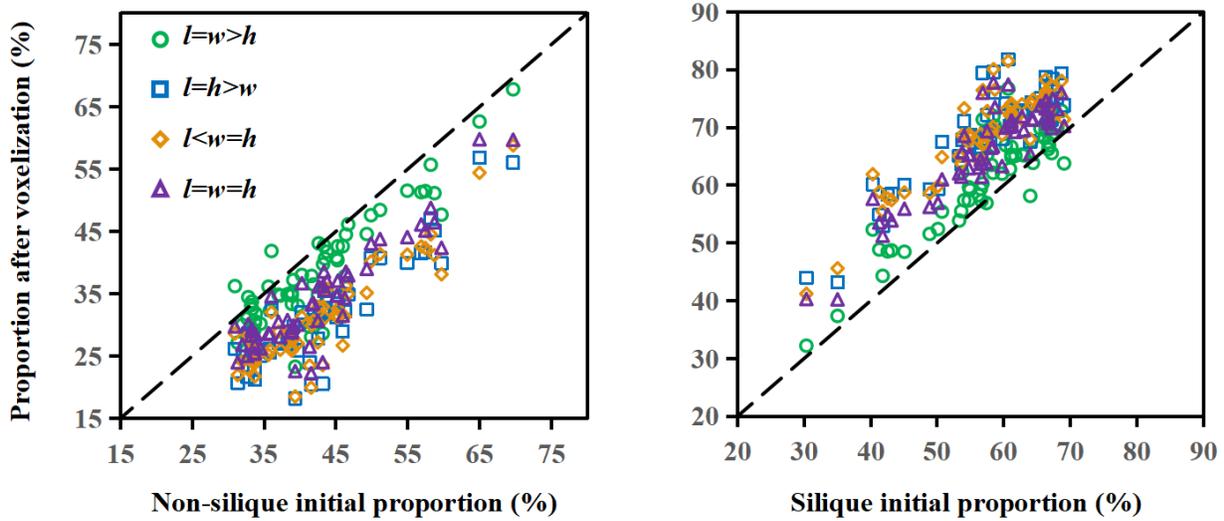

**Fig. 9.** Distribution of the proportion of two classes after dynamic voxelization with different voxel size. Two different voxel sizes (*length* (*l*), *width* (*w*), *height* (*h*)) are used to form four situation: 0.6cm, 0.6cm, 0.25 cm for *l=w>h*, *l=h>w*, *l<w=h* and 0.45cm×0.45cm×0.45 cm for *l=w=h*. (The voxel size values are hyper-parameters and chosen through network validation).

## 5.3 PST on three main structure types of rapeseed plants

In this study, the eight rapeseed plants cultivars mainly turn into three different structure types at the podding stage, and they are divided into S1, S2, and S3 as in Sec. 2.1. To compare the performance of PST on different structure types, we test the PST on three types separately. (Table 7). PST achieves the best performance on S1 with 95.08%, 97.59%, and 96.93% in mean IoU, overall accuracy, and mean F1-score, outperforming the results on S3 by 4.97, 2.34, and 3.74



percent points.

The reported results align with our expectation that the performance of PST on S1, S2, and S3 decreases. The structure complexity of S2 and S3 outweighs S1 since they have more branches. Thus, PST performs better on S1 with relatively simple structures. For S3, its first tiller location is at the bottom of the main stem, and there are more branches tillering from other branches than S2 has. With a more complex branch growth, the performance of PST on S3 is hence worse than that on S2. Overall, PST still ensures acceptable segmentation results with good precision. The three main structure types represent a wide range of rapeseed plant cultivars, proving that PST so far has a good capability to handle different rapeseed plant structures. Meanwhile, developing datasets from different plant species with various structures could further improve the generalization and performance of PST.

**Table 7**

The performance of PST on three types in the validation and testing samples. We report mean IoU overall accuracy, and mean F1-score to evaluate PST.

| Type | mIoU (%) | oAcc (%) | mF1 (%) |
| --- | --- | --- | --- |
| S1 | 95.08 | 97.59 | 96.93 |
| S2 | 93.49 | 96.84 | 95.78 |
| S3 | 90.11 | 95.25 | 93.19 |



## 5.4 Qualitative analysis of misclassified silique instance predictions

There are two main mistakes in the final outputs of silique instance predictions: (i) nearby different silique may be seen as one; (ii) part of the branch (i.e., non-silique) instances may be seen as siliques. We visualize the original coordinates and shifted coordinates in Fig. 10. It appears that the main reasons are from both the offset prediction branch and the semantic branch (i.e., PST). In offset prediction, the complex distribution of slim siliques makes the network hard to regress every silique to its respective centroid. Nearby siliques with boundary overlap may shift to the point between them. In the semantic branch, the wrongly predicted semantic labels from PST make the network apply coordinates shift on non-silique points, resulting in mistaken candidate clusters in both the original coordinate space and shifted coordinate space.

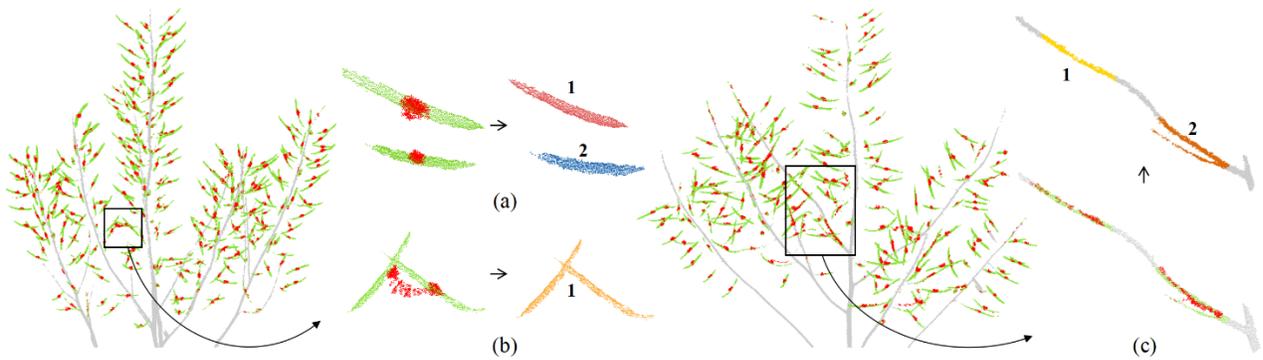

**Fig. 10.** Visualization of shifted coordinates set (red) and original coordinates set (green). (a) is the example of well-shifted siliques being predicted correctly. (b) shows nearby siliques shift to one centroid and are predicted as one. (c) denotes the mis-shift on the wrong semantic labels, resulting in branch instance being predicted as siliques.



# 6. Conclusion

The conventional point cloud segmentation based on the hard voxelization or down-sampling strategy limits the 3D phenotyping to simple plant samples, which is not suitable for segmenting dense plant point clouds with complex structures. To deal with this challenging problem, we chose the rapeseed plant point clouds acquired by HLS as a typical representation, whose morphology traits are complex, and the contextual information is highly susceptible to the density of the points. To segment HLS rapeseed plant point clouds, we adopted the dynamic voxelization and self-attention mechanism to realize per point feature learning without deteriorating the spatial information of the raw inputs. The proposed networks PST directly learned from the raw inputs with high spatial resolution and achieved an excellent trade-off between the segmentation results and the inference time. Moreover, PST can feasibly integrate with other two-stage networks as the semantic segmentation head to provide discriminative semantic labels for subsequent processes. Our results show that PST/PST-PG outperforms the state-of-the-art counterparts in the semantic and instance segmentation of HLS rapeseed plant point clouds. Without elaborated optimization, PST achieved the mean IoU of 93.96%, mean precision of 97.29%, mean recall of 96.52%, mean F1-score of 96.88%, and overall accuracy of 97.07% with an average inference time of 156.2ms per point cloud. PST-PG achieved 89.51%, 89.85%, 88.83% and 82.53% in mCov. mWCov, mPerc$_{90}$, and mRec$_{90}$, respectively. In addition, PST-PG obtains RMSE of 21.09 when IoU>75% and 25.87 when IoU>90%.



The proposed approach provides a new way to design the network for plant point cloud segmentation that is promising for high-precise 3D phenotyping of plants with complicated structures. Specifically, PST has shown a great ability to segment rapeseed plants with three different structure types. Future work should be performed on different plant species with various morphological properties to enrich the dataset of 3D plant point clouds and further generalize PST.

## Acknowledgement


This work was supported by the National Key R&D Program of China (2021YFD2000104), Key R&D Program of Zhejiang Province (2021C02057), Fundamental Research Funds for the Central Universities (226-2022-00217), and Zhejiang University Global Partnership Fund (188170 + 194452208/005).